\documentclass{article}

\PassOptionsToPackage{square,numbers,sort&compress}{natbib}
\usepackage[final]{neurips_2023} 

\usepackage{floatrow}
\usepackage[utf8]{inputenc} 
\usepackage[T1]{fontenc}    
\usepackage{hyperref}       
\hypersetup{
    colorlinks,
    linkcolor={red!50!black},
    citecolor={blue!50!black},
    urlcolor={blue!80!black}
}
\usepackage[dvipsnames]{xcolor}
\usepackage{url}            
\usepackage{booktabs}       
\usepackage{amsfonts}       
\usepackage{amsmath}
\usepackage{nicefrac}       
\usepackage{microtype}      
\usepackage{graphicx}
\usepackage{cancel}
\usepackage{color,soul}
\usepackage{mathtools}
\usepackage{wrapfig}
\usepackage{amssymb}
\usepackage{pgf}
\usepackage{array}
\usepackage{placeins}
\usepackage{tabularx}
\usepackage{makecell}
\usepackage[ruled,vlined]{algorithm2e}
\usepackage{bbold}
\usepackage{amsthm}
\usepackage{caption}
\usepackage{subcaption}
\usepackage{paralist}
\usepackage{multirow}
\usepackage{booktabs}
\usepackage{minitoc}
\usepackage{enumitem}
\usepackage{bm}
\usepackage[disable]{todonotes}
\usepackage{empheq}

\definecolor{fomo-green}{HTML}{70AD47}
\definecolor{fomo-purple}{HTML}{7030A0}
\definecolor{fomo-red}{HTML}{C00000}
\definecolor{fomo-blue}{HTML}{4472C4}
\definecolor{fomo-orange}{HTML}{ED7D31}

\newcommand{\object}{$\mathtt{object}\,$}
\newcommand{\characteristics}{$\mathtt{chars}\,$}
\newcommand{\constraint}{$\mathtt{constraint}\,$}
\newcommand{\container}{$\mathtt{container}\,$}
\newcommand{\rotangle}{$\mathtt{angle}\,$}
\newcommand{\orientation}{$\mathtt{direction}\,$}
\newcommand{\instruction}{\textcolor{fomo-blue}{$\mathtt{instruction}$} }
\newcommand{\positive}{\textcolor{fomo-green}{$\mathtt{positive}$} }
\newcommand{\negative}{\textcolor{fomo-orange}{$\mathtt{negative}$} }
\newcommand{\interface}{\textcolor{fomo-red}{$\mathtt{interface}$} }
\newcommand{\ctx}{\textcolor{fomo-purple}{$\mathtt{ctx}$} }
\newcommand{\ctxp}{\textcolor{fomo-green}{$\mathtt{ctx_{P}}$} }
\newcommand{\ctxn}{\textcolor{fomo-orange}{$\mathtt{ctx_N}$} }
\newcommand{\varinstr}{\textcolor{fomo-blue}{\mathtt{I}} }
\newcommand{\rtrue}{\textcolor{fomo-green}{$\mathtt{True}$} }
\newcommand{\rfalse}{\textcolor{fomo-orange}{$\mathtt{False}$} }
\newcommand{\mtrue}{\textcolor{fomo-green}{\mathtt{True}} }
\newcommand{\mfalse}{\textcolor{fomo-orange}{\mathtt{False}} }

\newtheorem*{theorem*}{Theorem}

\title{FoMo rewards: Can we cast foundation models\\as reward functions?}

\author{%
Ekdeep Singh Lubana$^{1,2,3}$\thanks{Work done during an internship at Qualcomm AI Research. Email: \texttt{eslubana@umich.edu}.}\;\qquad
Johann Brehmer$^1$\thanks{Order determined through \emph{Iene miene mutte}.}\qquad
Pim de Haan$^{1\dagger}$\qquad
Taco Cohen$^1$\\
$^1$Qualcomm AI Research\thanks{Qualcomm AI Research is an initiative of Qualcomm Technologies, Inc.}\;\qquad
$^2$University of Michigan\qquad
$^3$CBS, Harvard University%
}

\begin{document}

\maketitle
\begin{abstract}
We explore the viability of casting foundation models as generic reward functions for reinforcement learning. To this end, we propose a simple pipeline that interfaces an off-the-shelf vision model with a large language model. Specifically, given a trajectory of observations, we infer the likelihood of an instruction describing the task that the user wants an agent to perform. We show that this generic likelihood function exhibits the characteristics ideally expected from a reward function: it associates high values with the desired behaviour and lower values for several similar, but incorrect policies. Overall, our work opens the possibility of designing open-ended agents for interactive tasks via foundation models.
\end{abstract}

\section{Introduction}
\label{sec:intro}

\begin{wrapfigure}{}{0.5\textwidth}
  \vspace{-20pt}
  \begin{center}
    \includegraphics[width=0.95\textwidth]{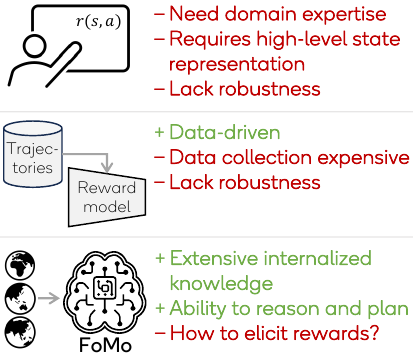}
  \end{center}
  \vspace{-10pt}
  \caption{
  \textbf{Reward functions in RL} can be constructed explicitly by humans (top), learned  from trajectory datasets (middle), or extracted from the world knowledge internalized by foundation models (bottom).
\vspace{-5pt}
}
\label{fig:reward_modeling}
\end{wrapfigure}

Recent advances in reinforcement learning (RL) algorithms have significantly simplified the design of data-driven, interactive agents~\cite{schulman2017proximal, schulman2015high, haarnoja2018soft, lillicrap2015continuous, mnih2013playing, silver2017mastering, anthony2017thinking, fujimoto2018addressing, o2016combining, nachum2017bridging, schulman2017equivalence, henderson2018deep}. Such algorithms generally rely on a reward function that captures a notion of ``desirable behavior'' to learn policies that output actions for engaging with the environment---which, itself, may~\cite{silver2017mastering, anthony2017thinking, wang2019benchmarking, hafner2020mastering, hafner2023mastering, hafner2019learning, lin2023learning} or may not be explicitly modeled~\cite{guez2019investigation, yarats2021improving, laskin2020curl, schwarzer2020data, schwarzer2021pretraining, laskin2021urlb}---in a manner that maximizes the reward. 
Though an integral component of the RL pipeline, the reward function is generally assumed to be provided by the user. 
However, designing a mathematically expressible reward function that can be optimized using RL algorithms is difficult for most real-world applications: e.g., even for the simple task of making a good cup of coffee, how does one mathematically express if a cup of coffee made by a robot is ``good''? 
%
Heuristically designed reward functions based on domain expertise can often allow \textit{hacking} in even fairly simple scenarios, i.e., they can enable learning of policies that maximize the reward, but do not engage in the desired behavior of interaction with the environment~\cite{hackingexamples, skalse2022defining, turner2020conservative, pan2022effects}. Research on \textit{reward modeling}~\cite{arora2021survey, ng2000algorithms, ziebart2008maximum, cao2021identifiability, ng1999policy, gupta2022unpacking, hu2020learning, szot2023bc} and \textit{imitation learning}~\cite{jiang2022efficient, janner2022planning, janner2021offline, chi2023diffusion, ajay2022conditional, chen2021decision, pearce2023imitating, hu2023instructed} has tried to address this situation, using offline trajectories of desirable behavior to either learn a reward function or to imitate behavior of the agent that produced the trajectories. Such frameworks however inherit the usual issues witnessed in learning-based systems; e.g., existence of shortcut solutions that lead to lack of robustness under domain shifts~\cite{de2019causal, skalse2023misspecification, skalse2023invariance, mckinney2023fragility, michaud2020understanding, gleave2020quantifying, tiencausal, everitt2021reward, ortega2021shaking, vuorio2022deconfounded}. Moreover, collecting offline trajectories makes these strategies expensive for several real-world applications; e.g., teaching a robot to not hurt a human by bumping into them is a difficult task to collect trajectories for. 

Foundation models (FoMos), especially large language models (LLMs), pretrained on huge, web-crawled datasets have revolutionized several fields of machine learning~\cite{radford2018improving, radford2019language, brown2020language, thoppilan2022lamda, touvron2023llama, touvron2023llama, hoffmann2022training, rae2021scaling, chowdhery2022palm, sanh2021multitask}. Demonstrating unprecedented capabilities for data-driven systems~\cite{bubeck2023sparks, wei2022emergent, rajkumar2022evaluating, bommasani2021opportunities}, the downstream adaptation of an off-the-shelf LLM has become the norm in most fields~\cite{wei2021finetuned, austin2021program, chen2021evaluating, raffel2020exploring, sanh2022multitask, jiang2023text2performer, zheng2023codegeex, nijkamp2022codegen, reid2022can}. In fact, recent work has demonstrated LLMs can be integrated with other modalities, such as vision, by training of minimalistic \textit{interface} components that transform the output of modality-specific models to a representation space ``legible'' to the LLM~\cite{hao2022language, merullo2022linearly, tsimpoukelli2021multimodal, alayrac2022flamingo, li2023blip}. Used alongside pretrained large vision models (LVMs), this interfacing pipeline further enables the use of LLMs in extremely broad applications that they were never trained for, e.g., text based image editing~\cite{kawar2022imagic, saharia2022photorealistic} or generation of 3D graphics~\cite{poole2022dreamfusion, lin2022magic3d}. Given the large amount of internalized knowledge in an LLM~\cite{petroni2019language, ding2023leveraging, alkhamissi2022review}, its ability to reason and plan (to an extent)~\cite{wei2022chain, kojima2022large, madaan2022language, huang2022large, bubeck2023sparks, qiao2022reasoning, dong2022survey, valmeekam2022large}, and a capability to ``perceive'' the real world via interfacing with perception models such as LVMs~\cite{merullo2022linearly, alayrac2022flamingo, yang2022empirical, zeng2022socratic, lyu2023macaw}, we argue foundation models are starting to elicit qualitative capabilities expected to be useful for design of generalist agents that can robustly function in open-ended scenarios.

Motivated by the above, in this work, we aim to assess if the framework of RL can benefit from the use of foundation models as well. Specifically focusing on the task of reward modeling, we make the following contributions.

\begin{itemize}[leftmargin=20pt, itemsep=1pt, topsep=3pt, parsep=3pt, partopsep=3pt]
    \item \textbf{FoMo as Rewards: A framework for eliciting reward functions from foundation models (Sec.~\ref{sec:pipeline}).} 
    Focusing on scenarios where the task of interest can be described via natural language, we propose a framework for casting decoder-based LLMs, the workhorse of SOTA language models, into reward functions. The assumption underlying our framework is that the LLM's input embedding space is ``approximately grounded''~\cite{merullo2022linearly, hao2022language, abdou2021can}, i.e., a pretrained LVM's outputs can be accurately processed by the LLM by learning a relatively light-weight interface model. Accordingly, our framework involves computing representations of observations from a trajectory via an LVM, using a \textit{learned interface} to morph these representations to the LLM's input embedding space, and evaluating the likelihood of the task's description using the LLM. 

    \item \textbf{Extensive qualitative analysis of FoMo as rewards (Sec.~\ref{sec:evals}).} 
    We perform an extensive evaluation to assess whether the rewards elicited by foundation models are sensible. Specifically, we use an oracle policy that engages in desired behavior and several adversarially perturbed versions thereof that, to different extents, differ in behavior from the desired one. For example, in a simple pick-and-place task, instead of picking the correct object, the agent might pick a different object from the scene. As we show, such adversarial policies achieve worse rewards than the correct one, indicating the use of FoMo Rewards is in fact viable for practical RL scenarios. 
    
\end{itemize}

Before proceeding, we emphasize that our work is currently focused on qualitative assessment of how to retrieve rewards from foundation models, i.e., we do not train RL policies from scratch and rely only on procedural environments to demonstrate that the use of foundation models for design of reward functions is in fact viable. We do believe evaluating RL policies trained using these rewards and demonstration of the pipeline on real world tasks, e.g., via use of an ego-centric dataset for interfacing LLMs and LVMs, is both feasible and important. We leave this analysis for future work.

\section{Related Work}
Several recent works have explored the use of foundation models for interactive tasks, generally focusing on learning policies based on behavior cloning~\cite{khandelwal2022simple, jiang2022vima, huang2023grounded, liu2022masked, stone2023open, brohan2022rt, driess2023palm, driess2022reinforcement, dai2023learning, zhang2022lad, nair2022learning, vemprala2023chatgpt, liu2022instruction, guhur2023instruction, hanjie2021grounding, lynch2020language, goyal2023language, bucker2022latte, schmeckpeper2020reinforcement, sermanet2016unsupervised}, pretraining representations for improved sample complexity~\cite{parisi2022unsurprising, 
 li2022pre, nair2022r3m, gadre2023cows, brandfonbrener2023inverse, liu2022masked, xiao2022masked, karamcheti2023language, hansen2021stabilizing, ze2023visual, hansen2022pre, liu2021aps, yuan2022pre, bahl2023affordances, liu2021behavior, baker2022video, radosavovic2023learning}, or for designing a world model for model-based RL~\cite{seo2023masked, seo2023multi, lin2023learning}. A few works have considered the task of reward modeling for training RL policies using foundation models~\cite{escontrela2023video, yu2023language, lin2022inferring, kwon2023reward}; we discuss approaches most relevant to our work below.

\paragraph{Rewards from foundation models.} Assuming access to task descriptions, prior work has considered the use of vision-language foundation models, e.g., CLIP~\cite{radford2021learning}, to infer the representational similarity between a task's description and the trajectory followed by an agent in the pursuit of performing that task~\cite{cui2022can, ma2023liv, ma2022vip, waytowich2019narration, goyal2019using, goyal2021zero, mahmoudieh2022zero}. Generally the vision component of these foundation models is trained on static images, which does not bode well for RL tasks where temporal information is important to account for. Hence, the works above retrain the vision pipeline from scratch on egocentric datasets such as Ego4D or on manually collected text-video pairs for the task of interest~\cite{fan2022minedojo, nair2022r3m}. A possible pitfall of such similarity based frameworks is that the training objective for CLIP-like pretraining has been shown by prior work to yield a ``bag-of-words'' model that is unable to capture concepts defined relatively between two objects; e.g., in an image of a person wearing a hat, the model enables recognizing the person and the hat, but not the fact that the hat is worn by the person~\cite{lewis2022does, yun2022vision, yuksekgonul2022and, thrush2022winoground}. As shown by Huang et al.~\cite{huang2023reminder}, this bag-of-words pitfall leads to a lack of ability to encode temporal constraints in such vision-language models. For example, the representation for instructions $\mathtt{Do\,A\,\,then\,B}$ and $\mathtt{Do\,B\,then\,\,A}$, where $\mathtt{A}$ and $\mathtt{B}$ are two specific tasks, will be extremely close. As expected, this can lead to undesirable behavior in the learned policies~\cite{huang2023reminder}.

\paragraph{Success detection.} Arguably most relevant to our work, Du et al.~\cite{du2023vision} investigate the use of multimodal foundation models based on a vision-interfaced decoder-only LLM, specifically Flamingo~\cite{alayrac2022flamingo}. Flamingo models are trained via learning of a cross-attention interface on top of a PaLM model~\cite{chowdhery2022palm}, yielding a highly performant baseline for multimodal tasks. Du et al. exploit Flamingo models and assess the successful completion of a task, i.e., provided a trajectory of visual observations from an agent executing a policy alongside the query ``has the task finished yet?'', does the model produce $\mathtt{True}$ or does it produce $\mathtt{False}$? Such a binary success token can be considered a reward function akin to commonly used success based rewards in learning board game agents~\cite{silver2017mastering, anthony2017thinking}. However, success based rewards can be extremely sparse, limiting their usability in several tasks of interest. In contrast, our proposed pipeline utilizes likelihood of the task description based on the trajectory up to the current instant, enabling a dense reward.

\section{Proposed framework: FoMo rewards}
\label{sec:pipeline}
\begin{figure}
\centering
        \includegraphics[width=\linewidth]{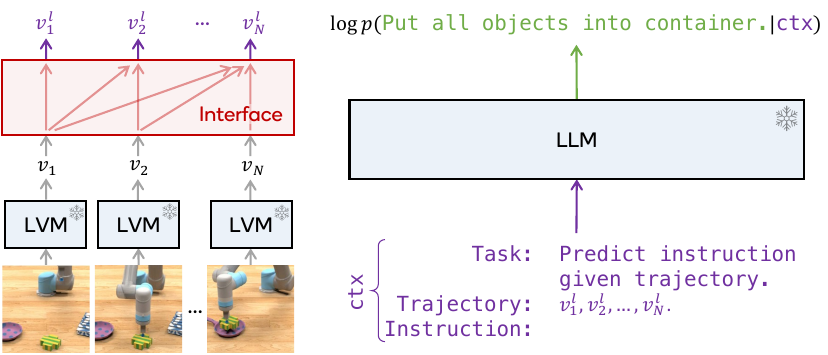}
        \caption{\textbf{Casting FoMo as Rewards.} Our proposed pipeline for casting pretrained foundation models involves learning an \textit{interface} model that maps representations of visual observations (\textcolor{gray}{$\{v_i\}$}) extracted via pretrained LVMs into a space ``legible'' to the LLM (\textcolor{Purple}{$\{v_i^{l}\}$}). This interface is a minimal, 1-layer Transformer in our experiments, allowing accommodation of temporal information. The remapped visual representations are added to a context (\ctx) inspired by instruction induction tasks for fine-tuning LLMs on downstream settings~\cite{honovich2022instruction}. Finally, the likelihood of an \textcolor{fomo-blue}{$\mathtt{instruction}$} describing the desired task is evaluated by inputting the context into a pretrained LLM via Eq.~\ref{eq:likelihood}.}
        \label{fig:pipeline}
\end{figure}

We next propose our framework for casting pretrained foundation models as reward functions (\textit{FoMo Rewards}), which involves assessing the likelihood of an instruction describing the task, provided representations of visual observations encompassing a trajectory of behavior (see Fig.~\ref{fig:pipeline} for an overview). We exploit the emergent grounding of LLMs, whereby a minimal transformation of a visual input's (or in fact other modalities' inputs) representation can be processed via an LLM to perform extremely involved tasks such as visual question answering~\cite{merullo2022linearly, abdou2021can}. Broadly, we perform the following steps.
\begin{enumerate}[leftmargin=20pt, itemsep=1pt, topsep=3pt, parsep=3pt, partopsep=3pt]
    \item \textbf{Infer visual representation of trajectory.} Given a set of frames from the agent's interaction with the environment up to the current instant $t$, we use an $\mathtt{LVM}$ to transform the frames into visual embeddings \textcolor{gray}{$\{v_1, v_2, \dots, v_t\}$}. 
    
    \item \textbf{Process visual representations via a pretrained interface model.} Trained using a dataset of instruction-trajectory data (training protocols discussed in Sec.~\ref{sec:interfacing}), the \interface transforms the visual representations to a space processable by the LLM, denoted \textcolor{Purple}{$\mathtt{Traj_t} = \{v^{l}_1, v^{l}_2, \dots, v^{l}_t\}$}. 
    We emphasize that prior works have trained interfaces as simple as a linear layer to enable language processing of visual scenes~\cite{merullo2022linearly}; however, these works focus on static image tasks, such as Visual Question Answering. We find incorporating temporal information can be difficult for a mere linear interface and address this by using a 1-layer decoder-only transformer.

    \item \textbf{Infer likelihood of task description given the trajectory's representation.} Finally, inspired by the ``instruction induction'' format of Honovich et al.~\cite{honovich2022instruction}, we embed the remapped visual representations into a broader context, denoted \textcolor{Purple}{$\mathtt{ctx_{t}}$} (see Fig.~\ref{fig:pipeline}), which is defined by (i) the phrase ``Task: Infer instruction given trajectory'', (ii) the interfaced visual trajectory's representation (\textcolor{Purple}{$\mathtt{Traj_t}$}), and (iii) a prompt ``Instruction:''. We then infer the log-likelihood of the instruction \textcolor{fomo-blue}{$\varinstr = \{i_1, i_2, \dots i_N\}$}, where \textcolor{fomo-blue}{$i_n$} denotes the $n^{\text{th}}$ token in the instruction, as follows.
    \begin{equation}
    \label{eq:likelihood}
        \log \mathcal{P}(\varinstr | \textcolor{Purple}{\mathtt{ctx_{t}}}) = \sum_n \log \mathcal{P}\left(\textcolor{fomo-blue}{\mathtt{i}_n} | \textcolor{Purple}{\mathtt{ctx_{t}}}, \textcolor{fomo-blue}{i_1, i_2, \dots, i_{n-1}}\right).
    \end{equation}
\end{enumerate}

The likelihood computed in Eq.~\ref{eq:likelihood} acts as our notion of reward. The intuition herein is that a grounded LLM can exploit its world knowledge to reason how likely is it that the agent is trying to engage in behavior described in the solve the task description provided by the user. For example, if the user says ``pick up the cup'' and the agent picks up a plate, ideally, the likelihood of the instruction should be lower in this scenario than the one in which the agent actually picks up the cup. This ideally implies if an agent were to engage in behavior described using the task description, the likelihood of the description would improve. Our experimental evaluation tests this intuition in extensive detail.

\subsection{Interfacing protocols}
\label{sec:interfacing}

\begin{figure}
\centering
        \includegraphics[width=\linewidth]{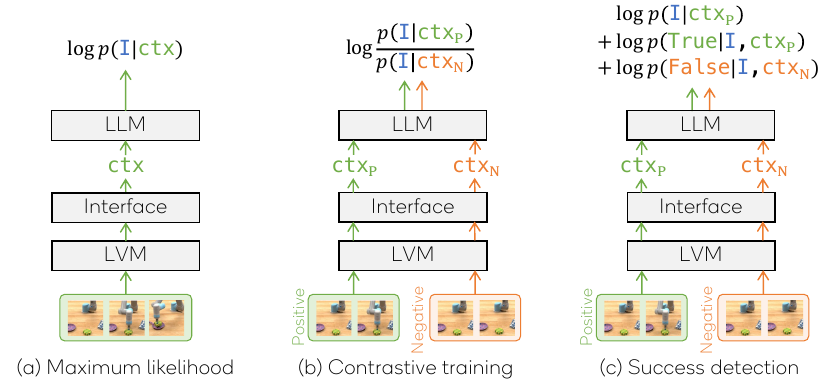}
        \caption{\textbf{Different protocols for training the interface models.} \textbf{(a) Maximum likelihood:} The interface is trained to map representations of the visual observations into a space such that the log-likelihood of the \instruction, given the trajectory, is maximized; this is the standard next word prediction loss decoder-based LLMs are trained with. \textbf{(b) Contrastive training:} A pair of \positive and \negative trajectories are defined, wherein the \positive trajectory does perform the desired task, while the \negative trajectory does not. The loss focuses on maximizing the ratio of the log-likelihood of the \instruction given the \positive trajectory with respect to the likelihood given the \negative trajectory. \textbf{(c) Success detection:} Similar to (b), \positive / \negative pairs are used; similar to (a), the likelihood of the \instruction is maximized only for the positive trajectory. An auxiliary objective is added, which predicts the success (\rtrue / \rfalse) of the positive/negative trajectory in solving the task specified by the \instruction.}
        \label{fig:interfaces}
\end{figure}

Unlike prior work that interfaces static images with an LLM to perform tasks such as Visual Question Answering, our goal is to interface a trajectory of observations that are temporally correlated. To this end, we use a 1-layer decoder-only Transformer model that involves sinusoidal positional encodings, a causal Attention module, and a 1-layer MLP to map representations from a VLM to an output with dimensions equal to that of the LLM's embeddings. In the benchmark we utilize in this paper (see Sec.~\ref{sec:setup}), the complexity of a task is often correlated with the number of actions required to solve it; correspondingly, the length of the trajectory can be used as a shortcut solution to output some paraphrase of the ground-truth instruction that specifies the task. It is easy to imagine such an issue emerging with realistic data involving, e.g., egocentric videos, where pick-and-place snippets may be smaller than movement from one place to another. We hence propose and evaluate three different protocols for training our interface models, as specified below (see Fig.~\ref{fig:interfaces} for an overview).

\textbf{Maximum likelihood:} The simplest of our protocols, herein we train the \interface to minimize the negative log-likelihood of instruction $\varinstr$ given a trajectory of observations that successfully solves the task. The mapped representations are inputted to the LLM in a manner similar to Fig.~\ref{fig:pipeline}, yielding the following objective.
    \begin{equation}
    \label{eq:direct}
        \mathcal{L}_{\mathtt{ML}} = - \log \mathcal{P}(\varinstr | \text{\ctx}).
    \end{equation}

\textbf{Contrastive training:} Herein, we first define a set of \positive and \negative trajectories that, respectively, correspond to an accurate versus inaccurate demonstration of the task specified by instruction \textcolor{fomo-blue}{$\mathtt{I}$}. To define a \negative trajectory, we (i) randomly choose and repeat an observation up to the length of the \positive trajectory or (ii) reverse the trajectory, preserving its length. The negative log-likelihood of \instruction is thus minimized for the \positive trajectory, but maximized for the \negative one. Since negative log-likelihood is unbounded, we clamp the loss on \negative trajectories up to an upper bound (set to $0.5$) in practice, yielding the following objective.  
    \begin{equation}
    \label{eq:contrast}
        \mathcal{L}_{\mathtt{Contrast}} = -  \left(\log \mathcal{P}(\varinstr | \text{\ctxp}) - \mathtt{clamp}\left(\log \mathcal{P}(\varinstr | \text{\ctxn}),\, 0.5 \right) \right),
    \end{equation}
    where \ctxp / \ctxn denote the instruction induction context (see Sec.~\ref{fig:pipeline}) with representations of the \positive / \negative trajectory substituted within it.

\begin{wrapfigure}[12]{}{0.44\textwidth}%
  \centering%
  \vskip-18pt%
  \includegraphics[width=\textwidth]{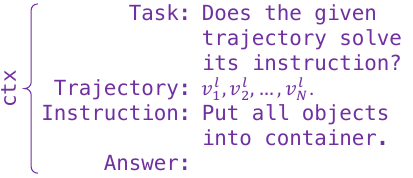}%
  \vskip-6pt%
  \caption{%
    \textbf{Context used for success detection.} The LLM is given the context above and its likelihood of outputting the tokens {\rtrue / \rfalse} is used to instantiate the loss in Eq.~\ref{eq:success}.%
  }%
\label{fig:sdet_context}%
\end{wrapfigure}

\textbf{Success detection:} Finally, we evaluate the use of an auxiliary loss that promotes a binary, True/False distinction between the \positive versus \negative trajectory alongside the negative log-likelihood maximization of the instruction \textcolor{fomo-blue}{$\mathtt{I}$} given the \positive trajectory. The primary benefit of this formalism is that it avoids training instabilities associated with maximizing the negative log-likelihood of the \negative trajectories in Eq.~\ref{eq:contrast}. However, the use of a decoder LLM makes instantiation of this framework in a standard manner of train a classifier on some representation infeasible. We thus cast this auxiliary task in the instruction following format of Wei et al.~\cite{wei2022finetuned} (see Fig.~\ref{fig:sdet_context}).
    \begin{equation}
    \label{eq:success}
        \mathcal{L}_{\mathtt{Success}} = - \left(\log \mathcal{P}(\varinstr | \text{\ctxp}) + \log \mathcal{P}( \mtrue| \varinstr, \text{\ctxp}) + \log \mathcal{P}( \mfalse| \varinstr, \text{\ctxn}) \right).
    \end{equation}

\section{Experimental protocols}
\label{sec:setup}

\begin{figure}%
\centering%
        \includegraphics[width=\linewidth]{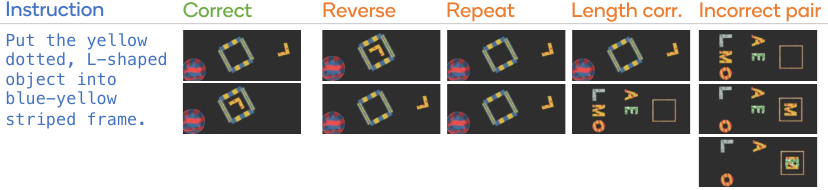}%
        \caption{\textbf{Perturbed trajectories.} We perturb an oracle policy's trajectory of observations in systematic ways to assess if our pipeline elicits desirable behaviors expected from a reward function. We specifically evaluate likelihood of instruction given the: (i) \textit{correct} trajectory; (ii) its \textit{reverse}; (iii) an action-less trajectory where a randomly sampled observation from the correct trajectory is repeated; and (iv) an entirely incorrect trajectory corresponding to another task or instruction.}%
        \label{fig:bad_trajs}%
\end{figure}

We next define the experimental setup used in this paper to evaluate the use of FoMo as rewards. As noted in Sec.~\ref{sec:intro}, our focus in this work is to perform a qualitative assessment of the question whether FoMos can serve as reward functions. Accordingly, we also describe several evaluation protocols aimed at systematically perturbing an oracle policy and assess the behavior of our pipeline under these systematic perturbations.

\paragraph{Evaluation Benchmark.} Our evaluation focuses on the recently proposed multimodal, goal-directed RL benchmark \textit{VIMABench}~\cite{jiang2022vima}. Specifically, VIMABench involves thirteen tabletop manipulation tasks that are specified via the use of a ``multimodal prompt''; e.g., the prompt may specify a pick-and-place task that requires an \object to be moved into a certain \container in the scene or a manipulation task that requires it to be rotated by a specific \rotangle. Herein, both the variables \object and \container are generally specified via an image. However, our proposed pipeline focuses on likelihood evaluation of the instruction that specifies a task in language, making VIMABench's multimodal prompts infeasible for our setup. To circumvent this, we write a preprocessing wrapper that transforms variables specified via images into text; e.g., an image of a blue-green polka-dotted flower will be replaced with the text \texttt{blue-green polka-dotted flower}. This yields us a unimodal version of the VIMABench benchmark. Note that only seven of VIMABench's tasks are feasible for easy conversion to such unimodal prompts (e.g., excluding when precise object coordinates are needed, which cannot be cheaply extracted without running an object detection module). For brevity, we often refer to this unimodal version of VIMA as \texttt{VIMA(Uni)}.

In \texttt{VIMA(Uni)}, objects (including containers) are denoted via a tuple of $\left(\mathtt{chars}, \mathtt{identifier}\right)$. The variable $\mathtt{identifer}$ takes values from a list of predefined object and container shapes, while the variable $\mathtt{chars}$ takes values from a list of predefined characteristics that includes colors, textures, and orientations. An example prompt and start/end frame from a top-view camera of a VIMA agent performing the task specified by the prompt are shown in Fig.~\ref{fig:bad_trajs}. The \texttt{VIMA(Uni)} tasks with their variable prompts are listed below. Here subscripts indicate different objects in the scene, while the variable \orientation~is one of the ordinal directions, \rotangle~is a multiple of 30 degrees, and \constraint~is an obstacle object with which interaction is to be avoided. 
\begin{itemize}[leftmargin=20pt, itemsep=1pt, topsep=3pt, parsep=3pt, partopsep=3pt]
    \item \textbf{Task 1: Same profile.} Put all objects with the same attributes as \characteristics~\container~into it.
    \item \textbf{Task 2: Manipulate old neighbor.} First put \characteristics$_1$~object into \container$_1$, then put the object that was previously at its \orientation~into the same container.
    \item \textbf{Task 3: Scene Understanding.} Put the \characteristics$_1$~object in scene into the \characteristics$_2$~object.
    \item \textbf{Task 4: Pick in order then restore.} Put \characteristics$_1$~object into the \characteristics$_2$~container and then the \characteristics$_3$~container. Finally restore it into its original container.
    \item \textbf{Task 5: Simple Manipulation.} Put the \characteristics$_1$~object into the \characteristics$_2$~container.
    \item \textbf{Task 6: Rotate.} Rotate \characteristics~\object~by~\rotangle~degrees.
   \item \textbf{Task 7: Sweep without exceeding.} Sweep \characteristics$_1$~\object~into the \characteristics$_2$~\container~without exceeding \constraint.
\end{itemize}

\subsection{Systematic perturbations for evaluation} 

To evaluate if the recasting of FoMo as rewards is sensible, we evaluate the likelihood of several systematically perturbed setups. Specifically, we evaluate the following two scenarios.
\paragraph{Perturbed Trajectories.} In this evaluation, we alter the trajectory taken by an oracle policy to perform the task specified by the \instruction in the following different manners (see Fig.~\ref{fig:bad_trajs} for an overview). We evaluate the likelihood of the \instruction given these perturbed trajectories.
        \begin{enumerate}[leftmargin=20pt, itemsep=1pt, topsep=3pt, parsep=3pt, partopsep=3pt]
            \item $\mathtt{PT_{Rev}}$: Reverse the oracle trajectory by rendering it from end to start.
            \item $\mathtt{PT_{Rep}}$: Randomly choose an observation (a frame) from the oracle trajectory and repeat it to match the oracle trajectory in length. 
            \item $\mathtt{PT_{Len}}$: Randomly choose an observation from the oracle trajectory and replace it with an observation from another trajectory that may or may not correspond to the same task.
            \item $\mathtt{PT_{Inc}}$: Use an entirely different oracle trajectory that may or may not correspond to the same task.
        \end{enumerate}

    Intuitively, the perturbations above are designed to assess if use of FoMo as rewards elicits desirable behaviors expected from a reward function. For example, $\mathtt{PT_{Rev}}$ helps us assess \textit{whether our pipeline assigns higher reward to appropriate temporal order of correct actions over just the presence of correct actions}. Indeed, in a pick and place task, the mere fact that the pick and place actions were performed may be sufficient to judge that the task involves picking a certain object and placing it in a specific place. However, an ideal reward function would assign higher reward to the correct temporal order, than the mere use of pick and place actions.

\paragraph{Perturbed Instructions.} Alter the instructions specifying the task such that identifying attributes of the objects that the task is supposed to be performed on are converted to attributes of other objects present in the scene. We specifically focus on the following perturbations (see Fig.~\ref{fig:bad_trajs} for an overview).
        \begin{enumerate}[leftmargin=20pt, itemsep=1pt, topsep=3pt, parsep=3pt, partopsep=3pt]
            \item $\mathtt{PI_{Obj}}$: Alter the value of \object and \container identifiers from target objects/containers to another set of objects/containers present in the scene.
            \item $\mathtt{PI_{Col}}$: Alter the \characteristics of target objects/containers by replacing their \textit{color} with the color of another object present in the scene. 
            \item $\mathtt{PI_{Tex}}$: Alter the \characteristics of target objects/containers by replacing their \textit{texture} with the texture of another object present in the scene.
            \item $\mathtt{PI_{Comb}}$: Combine all alterations listed above into a single altered instruction.
        \end{enumerate}

    The motivations behind defining the above perturbed instructions is similar to that of the perturbed trajectories. For example, $\mathtt{PI_{Obj}}$ alters the instruction such that the task to be performed remains the same, e.g., still a pick and place task, but the specific objects that are supposed to be acted upon are changed. This implies the objects the oracle policy acts upon, i.e., the objects specified in the original, unaltered instruction, will be different from the objects this perturbed instruction specifies. Assigning a lower reward to this perturbed instruction--trajectory pair, when compared to the correct one, indicates \emph{our pipeline yields a reward function that accounts for the precise physical specification, i.e., the actions being performed and what objects they are performed on.}

\begin{figure}%
\centering%
        \includegraphics[width=\linewidth]{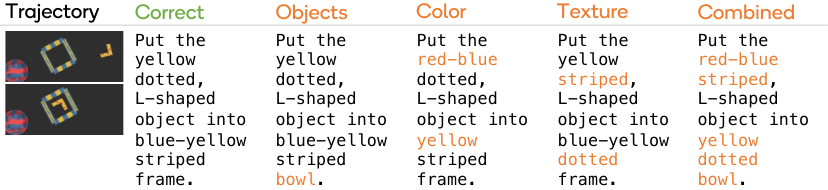}%
        \caption{\textbf{Perturbed Instructions.} We perturb the instruction corresponding to an oracle trajectory in systematic ways to assess if our pipeline elicits desirable behaviors expected from a reward function. Given the trajectory, we specifically evaluate likelihood of the following perturbed instructions: (i) \textit{Object}: objects/containers are altered to other objects/containers in the scene; (ii) \textit{Color}: the color specification of the target objects/containers is altered to other colors present in the scene; (iii) \textit{Texture}: the texture specification of the target objects/containers is altered to other textures present in the scene; and (iv) \textit{Combined}: all alteration are combined into a single, altered instructions.}%
        \label{fig:bad_instrs}%
\end{figure}

\section{Results: FoMo is rewarding}
\label{sec:evals}

\begin{table*}
    \setlength{\tabcolsep}{3.2pt}
    \caption{\label{tab:results_trajs}\textbf{Average log likelihood of \instruction given correct versus perturbed trajectories (higher is better).} We compute the likelihood of the instruction specifying the task given several alterations of the correct trajectory, as specified in Sec.~\ref{sec:setup}. The average is computed over all trajectories corresponding to a task ($\sim$4300 per task). Broadly, the results show the use of FoMo as rewards is indeed inline with desired behavior when the \interface is trained via the contrastive or success detection protocol: for the correct trajectories, the average likelihood of the instruction is generally much higher than the altered variants. For a subset of the tasks (e.g., T6 or Rotate), the pipeline struggles with assigning noticeably higher reward to the correct trajectory compared with the reverse or repeated one.\vspace{-8pt}}
    \centering
    \scriptsize
    \begin{tabular}{@{}c|rrrrr|rrrrr|rrrrr@{}}
        \toprule
         & \multicolumn{5}{c|}{Maximum likelihood}                & \multicolumn{5}{c|}{Contrastive}                    & \multicolumn{5}{c}{Success detection}               \\ \midrule
        Task & Correct    & $\mathtt{PT_{Rev}}$ & $\mathtt{PT_{Rep}}$ & $\mathtt{PT_{Len}}$ & $\mathtt{PT_{Inc}}$ & Correct    & $\mathtt{PT_{Rev}}$ & $\mathtt{PT_{Rep}}$ & $\mathtt{PT_{Len}}$ & $\mathtt{PT_{Inc}}$ & Correct    & $\mathtt{PT_{Rev}}$ & $\mathtt{PT_{Rep}}$ & $\mathtt{PT_{Len}}$ & $\mathtt{PT_{Inc}}$ \\ \midrule
        T1 & \textbf{-0.80} & -1.18 & -1.14 & -7.03 & -37.49 & \textbf{-1.52} & -77.28 & -44.30 & -25.49 & -49.51 & \textbf{-1.25} & -2.97 & -2.28 & -13.07 & -39.51 \\
        T2 & \textbf{-3.60} & -5.95 & -6.59 & -18.19 & -63.92 & \textbf{-6.42} & -93.89 & -64.60 & -38.71 & -69.44 & \textbf{-7.14} & -13.98 & -10.92 & -25.80 & -77.56 \\
        T3 & \textbf{-1.79} & -2.32 & -2.59 & -9.61 & -33.85 & \textbf{-3.11} & -45.90 & -22.52 & -15.99 & -28.89 & \textbf{-3.18} & -4.40 & -3.80 & -10.80 & -38.85 \\
        T4 & \textbf{-6.22} & -6.33 & -7.53 & -26.06 & -87.25 & \textbf{-9.71} & -9.83 & -10.08 & -43.20 & -94.69 & \textbf{-9.84} & -9.97 & -10.04 & -25.27 & -101.98 \\
        T5 & \textbf{-4.56} & -5.72 & -7.16 & -15.21 & -37.85 & \textbf{-6.55} & -32.30 & -19.33 & -19.21 & -34.81 & \textbf{-6.84} & -9.44 & -8.58 & -13.06 & -35.60 \\
        T6 & -4.10 & \textbf{-4.08} & -4.12 & -11.24 & -34.83 & -4.65 & \textbf{-4.61} & -4.68 & -10.26 & -33.98 & -4.92 & \textbf{-4.86} & -4.88 & -11.76 & -32.94 \\
        T7 & \textbf{-1.84} & -2.22 & -2.72 & -21.45 & -99.06 & \textbf{-4.38} & -19.04 & -11.53 & -45.36 & -129.28 & \textbf{-4.07} & -5.05 & -4.89 & -52.59 & -147.13 \\
        \bottomrule
    \end{tabular}
\end{table*}

\begin{table*}
    \setlength{\tabcolsep}{4pt}
    \caption{\label{tab:results_instrs}\textbf{Average log likelihood of perturbed instructions given the correct trajectory (higher is better).} We compute the likelihood of instructions altered in the manner specified in Sec.~\ref{sec:setup}, while using the oracle trajectory corresponding to the unaltered \instruction. The average is computed over all trajectories corresponding to a task ($\sim$4300 per task). Broadly, the results show the use of FoMo as rewards is indeed inline with desired behavior:
    for all tasks, only for the correctly specified instructions, the average likelihood of the instruction is higher than the altered variants. We show the highest likelihoods in italics, but note that we are comparing likelihoods of texts of slightly varying lengths (generally the difference is 2--4 tokens).\vspace{-8pt}}
    \centering
    \scriptsize
    \begin{tabular}{@{}c|rrrrr|rrrrr|rrrrr@{}}
        \toprule
         & \multicolumn{5}{c|}{Maximum likelihood}                & \multicolumn{5}{c|}{Contrastive}                    & \multicolumn{5}{c}{Success detection}               \\ \midrule
        Task & GT    & $\mathtt{PI_{Obj}}$ & $\mathtt{PI_{Col}}$ & $\mathtt{PI_{Tex}}$ & $\mathtt{PI_{Comb}}$ & GT    & $\mathtt{PI_{Obj}}$ & $\mathtt{PI_{Col}}$ & $\mathtt{PI_{Tex}}$ & $\mathtt{PI_{Comb}}$ & GT    & $\mathtt{PI_{Obj}}$ & $\mathtt{PI_{Col}}$ & $\mathtt{PI_{Tex}}$ & $\mathtt{PI_{Comb}}$ \\ \midrule
        T1 & \emph{-0.81} & -12.96 & -4.17 & -12.07 & -30.06 & \emph{-1.49} & -12.34 & -4.23 & -11.24 & -26.72 & \emph{-1.24} & -12.25 & -4.11 & -9.92 & -27.00 \\
        T2 & \emph{-3.61} & -18.27 & -8.60 & -25.72 & -47.75 & \emph{-6.43} & -21.87 & -11.09 & -23.86 & -44.28 & \emph{-7.15} & -20.81 & -11.07 & -22.95 & -38.51 \\
        T3 & \emph{-1.84} & \emph{-1.84} & -7.36 & -27.14 & -31.58 & \emph{-3.10} & \emph{-3.10} & -8.08 & -24.15 & -27.66 & \emph{-3.23} & \emph{-3.23} & -8.08 & -22.98 & -26.49 \\
        T4 & \emph{-6.18} & -36.17 & -14.20 & -31.10 & -68.44 & \emph{-9.71} & -40.91 & -16.05 & -28.34 & -63.82 & \emph{-9.87} & -37.82 & -16.11 & -27.19 & -59.78 \\
        T5 & \emph{-4.50} & -17.48 & -8.91 & -28.79 & -48.45 & \emph{-6.53} & -18.54 & -10.47 & -25.18 & -42.91 & \emph{-6.81} & -18.35 & -10.62 & -24.85 & -42.25 \\
        T6 & -4.10 & \emph{-3.99} & -6.38 & -12.97 & -23.88 & \emph{-4.66} & \emph{-4.67} & -6.64 & -12.43 & -20.00 & -4.92 & \emph{-4.85} & -7.02 & -12.89 & -20.05 \\
        T7 & \emph{-1.85} & -41.31 & -10.76 & -42.59 & -86.14 & \emph{-4.38} & -39.83 & -11.47 & -35.48 & -72.73 & \emph{-4.07} & -42.81 & -11.25 & -36.05 & -78.31 \\
        \bottomrule
    \end{tabular}
\end{table*}

With our experimental setup and evaluation protocols set up, we are now ready to answer our motivating question: ``Can foundation models be cast as reward functions?'' We instantiate our framework (see Sec.~\ref{sec:pipeline}) by using an off-the-shelf, $300$M parameter ViT from the PyTorch image models library hosted on HuggingFace~\cite{rw2019timm}. The model is trained on the ImageNet-21K dataset using the protocol proposed by Steiner et al.~\cite{steiner2021train}. We use the recently released Mosaic pretrained transformers (MPT)~\cite{mpt} as the LLM workhorse in our pipeline, focusing on the $1$B parameter model that was pretrained using the RedPajama dataset~\cite{together2023redpajama} and further instruction fine-tuned using the Databricks Dolly dataset~\cite{DatabricksBlog2023DollyV2}. We use approximately $270$K offline trajectories from an oracle policy from VIMA(Uni) to train our \interface models and $30$K trajectories to evaluate the overall reward function it yields (trajectories are approximately uniformly distributed across tasks). Training occurs at a batch-size of 8 for maximum likelihood and batch-size of 4 for contrastive and success detection protocols. Note the LVM and LLM remain frozen and their parameters are not changed at all. The oracle utilizes privileged state information to perform the specified task, though it can at times fail due to imperfect inverse kinematics by the underlying PyBullet engine.

\begin{figure}
\centering
\begin{subfigure}[b]{\textwidth}
  \centering
  \centerline{\includegraphics[width=\columnwidth]{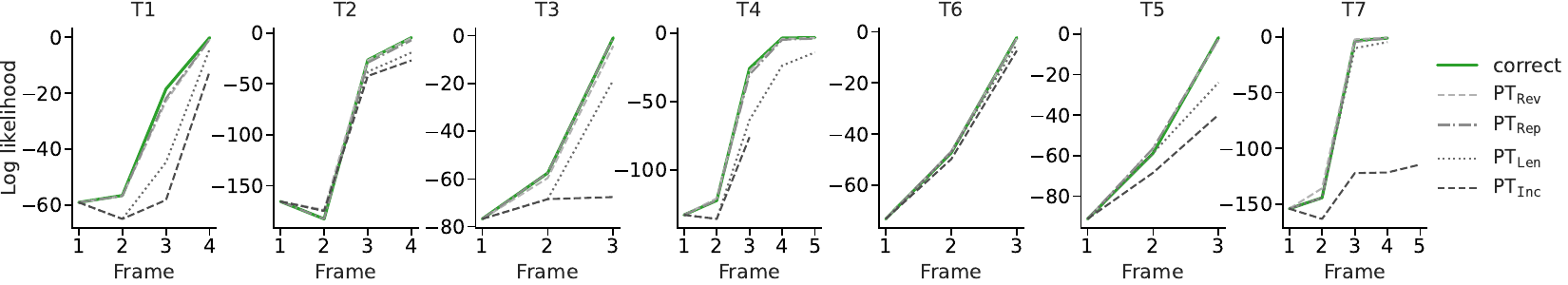}}
  \label{fig:intent_llhood_dm}
  \caption{Maximum likelihood training.}
\end{subfigure}
\begin{subfigure}[b]{\textwidth}
  \centering
  \centerline{\includegraphics[width=\columnwidth]{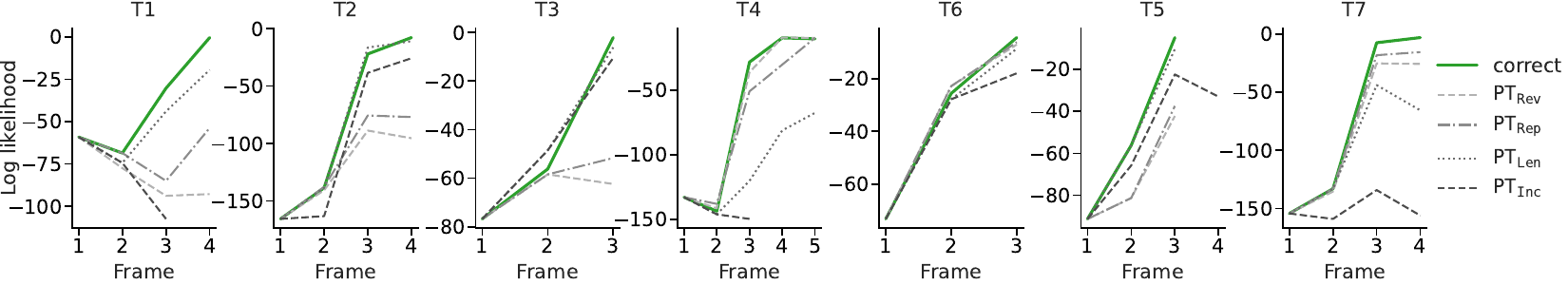}}
  \label{fig:intent_llhood_contrast}
  \caption{Contrastive training.}
\end{subfigure}
\begin{subfigure}[b]{\textwidth}
  \centering
  \centerline{\includegraphics[width=\columnwidth]{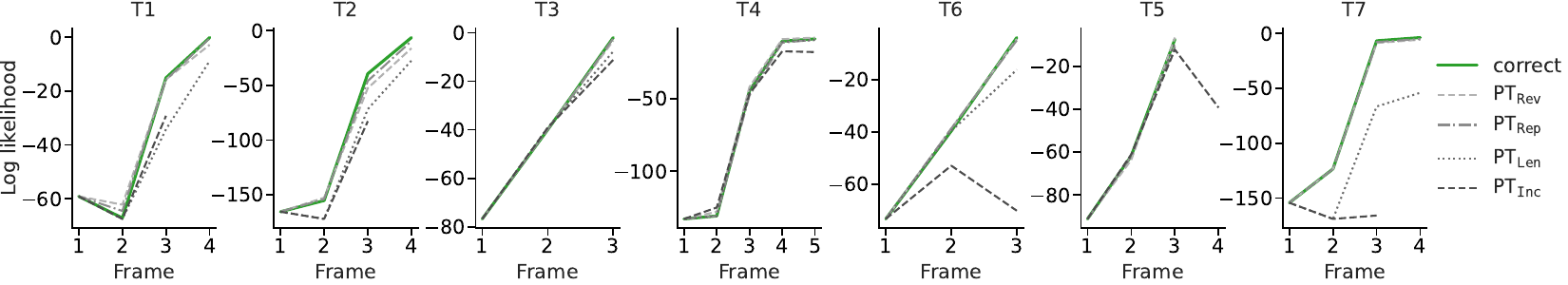}}
  \label{fig:intent_llhood_sdet}
  \caption{Success detection.}
\end{subfigure}
\caption{\textbf{FoMo rewards when training with different training protocols.} We plot \instruction log-likelihoods as a function of actions taken in the environment in 7 trajectories, showing results for the different training protocols proposed in Sec.~\ref{sec:interfacing}. On most tasks and at most steps, the correct policy achieves higher rewards than the perturbed behaviors when the interface is trained via the contrastive or success detection training protocols. However, mere maximum likelihood training, the simplest of our protocols, clearly suffers from shortcut solutions and yields similar results for both correct and perturbed trajectories.
\label{fig:intent_llhoods}
}
\end{figure}

We report the following sets of results: (i) likelihood of \instruction given the oracle versus perturbed trajectories, averaged over all instruction--trajectory pairs of a task (Tab.~\ref{tab:results_trajs}); (ii) likelihood of correct versus perturbed \instruction given the oracle trajectory for the correct instruction, averaged over all instruction--trajectory pairs of a task (Tab.~\ref{tab:results_instrs}); and (iii) progress of the likelihood of the correct instruction given oracle versus perturbed trajectories is plotted as a function ofobservations seen after the execution of an action (Fig.~\ref{fig:intent_llhoods}), an evaluation inspired by the “intentscoring” experiment by Karamcheti et al.~\cite{karamcheti2023language}.

We find that across almost all tasks, the correct behaviours achieve higher likelihoods than all perturbed trajectories when the contrastive and success detection training protocols are used. The maximum likelihood objective, our simplest protocol, however clearly suffers from shortcut solutions: indeed, in Tab.~\ref{tab:results_trajs} and Fig.~\ref{fig:intent_llhoods}, we see this protocol yields similar likelihoods for both correct trajectories and the perturbed ones (e.g., repeat frames yield very similar results). These results provide evidence that while FoMo rewards are viable and that agents trained on them may learn the intended behaviours, the training protocol must be devised with caution depending on the task and setup.

\section{Future work}
\label{sec:future}

In this work, we focused on demonstrating the viability of using foundation models as reward functions by comparing the rewards achieved from various policies.
Motivated by the promising results, the logical next step is to train agents on these reward functions with off-the-shelf RL algorithms. The VIMA environment used in this study provided a practical benchmark, but is not very representative of the real-world environments. Future work should study richer environments and training on large egocentric datasets. Finally, the use of foundation models to define desirable behaviors opens up the possibility of exploiting in-context learning as a means of adopting behaviours out of domain.

\subsection*{Acknowledgements}

We would like to thank Pietro Mazzaglia and Risto Vuorio for fruitful discussions.

\clearpage
\bibliography{fomobib}

\end{document}